\documentclass[10pt,twocolumn,letterpaper]{article}

\usepackage{iccv}
\usepackage{multirow}
\usepackage{times}
\usepackage{epsfig}
\usepackage{graphicx}
\usepackage{amsmath}
\usepackage{amssymb}
\usepackage{authblk}
\usepackage[accsupp]{axessibility}  

\usepackage[pagebackref=true,breaklinks=true,letterpaper=true,colorlinks,bookmarks=false]{hyperref}

\usepackage{adjustbox}

\iccvfinalcopy 

\def\iccvPaperID{0004} 

\ificcvfinal\fi

\begin{document}

\title{Scene Designer:  a Unified Model for Scene Search and Synthesis from Sketch}


\author[1]{Leo Sampaio Ferraz Ribeiro}
\author[2]{Tu Bui}
\author[2, 3]{John Collomosse}
\author[1]{Moacir Ponti}

\affil[1]{ICMC, Universidade de S{\~a}o Paulo -- S{\~a}o Carlos/SP, Brazil \authorcr
  \{\tt leo.sampaio.ferraz.ribeiro,ponti\}@usp.br}
\affil[2]{CVSSP, University of Surrey -- Guildford, Surrey, UK \authorcr
  \{\tt t.bui,j.collomosse\}@surrey.ac.uk}
\affil[3]{Adobe Research, Creative Intelligence Lab -- San Jose, CA, USA \authorcr \tt collomos@adobe.com}

\maketitle
\begin{abstract}
{Scene Designer is a novel method for searching and generating images using free-hand sketches of scene compositions; i.e. drawings that describe both the appearance and relative positions of objects.  Our core contribution is  a single unified model to learn both a cross-modal search embedding for matching sketched compositions to images, and an object embedding for layout synthesis. We show that a graph neural network (GNN) followed by Transformer under our novel contrastive learning setting is required to allow learning correlations between object type, appearance and arrangement, driving a mask generation module that synthesizes coherent scene layouts, whilst also delivering state of the art sketch based visual search of scenes.}  
\end{abstract}

\section{Introduction}

Creativity is increasingly inspired by the wealth of visual content online.   Visual search eases content discovery and re-use, whilst generative artwork is emerging as a novel genre, driven by models trained on thousands of images.  Yet the {\em fusion of search and synthesis} is under-explored.  Generative content oveoffers users control but rarely the quality and diversity of real images. By contrast, search offers quality but not customization.  Recent works have explored both image search and generation guided by free-hand sketches; an intuitive way to communicate visual intent.  Sketch therefore offers an opportunity to unify search and synthesis technologies within a creative workflow.

This paper presents Scene Designer; a unified model for searching and generating scenes using free-hand sketches of scene compositions; i.e. drawings that describe both the appearance and relative positions of multiple objects.  Our model provides a cross-modal search embedding  where similarity of visual compositions comprising sketches and images (or mixtures of both) can be measured, as well as an object-level embedding for the synthesis of scene layouts to generate images.  Our technical contributions are:

\begin{figure}[t!]
    \centering
    \includegraphics[width=\linewidth]{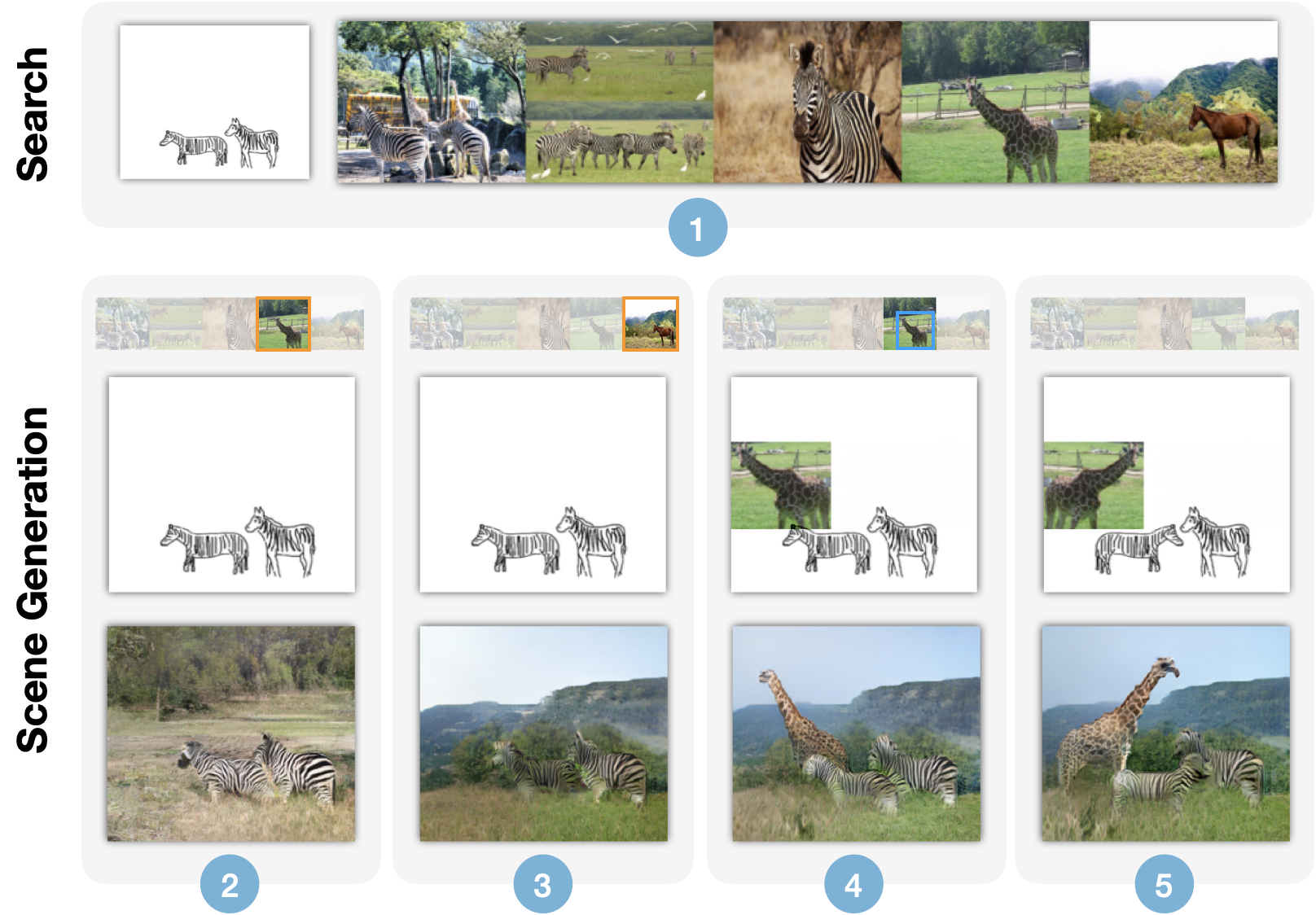}
    \caption{Scene Designer enables iterative design of image compositions through retrieval and synthesis.  Row 1: User sketches initial concept, and matching scenes are returned (1). Row 2: In each column, the user uses the search results to help compose the final image; an orange square means the background was selected (2, 3) while the blue one means an object crop was chosen and added to the composition (4); on the final stage the user poses the objects as desired, and the final scene is synthesized (5).}
    \label{fig:teaser}
\end{figure}

\noindent {\bf 1. Compositional Sketch Search.} We propose a hybrid graph neural network (GNN) and Transformer architecture to learn a metric search embedding for comparing sketched and photographic scenes. Existing sketch based image retrieval (SBIR) methods predominantly match queries containing a single, dominant object invariant to its position within an image. Our novel contrastive training matches sketched compositions containing multiple objects.

\begin{figure*}[t!]
    \centering
    \includegraphics[width=\linewidth]{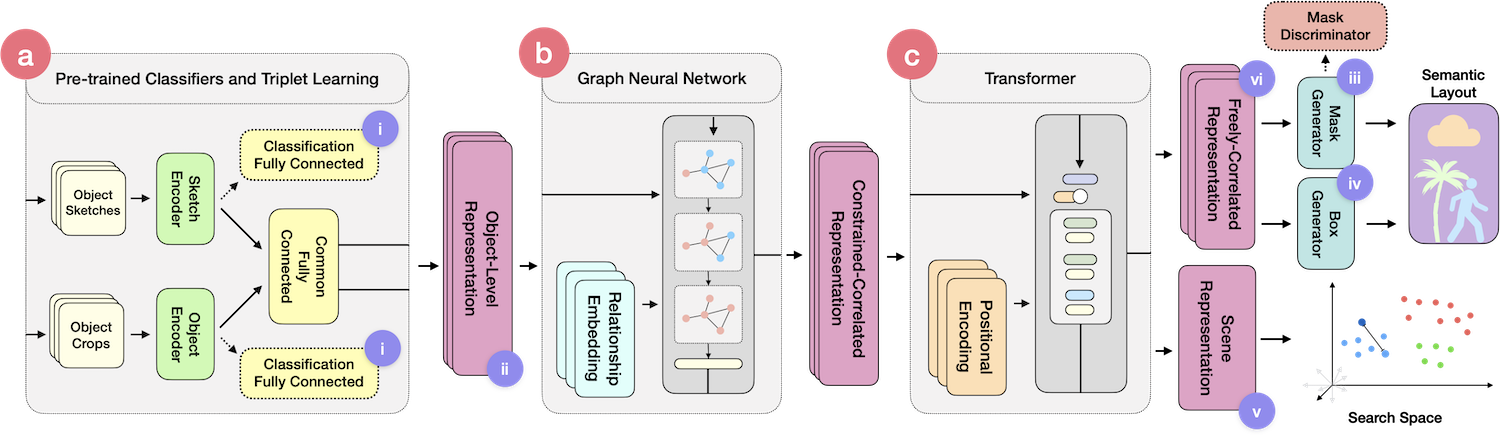}
    \caption{Proposed architecture of Scene Designer.  Compositions (spatial arrangements of sketched or photoreal objects) are encoded via (a) to produce an \textit{object-level representation}; \ie a feature encoding for each object in the scene. A scene graph is constructed from these and encoded via GNN (b) to yield the  \textit{constrained-correlated representation}, and subsequently via Transformer to produce (c) the \textit{freely-correlated representation} and \textit{scene representation}. The roman numerals indicate where each of the losses  (see Sec. \ref{sec:olr}, \ref{sec:syn} and \ref{sec:sbir}) are applied during training: (i) Cross-entropy classification $\mathcal{L}_{CCE}$, (ii) Triplet loss $\mathcal{L}_{tri}$, (iii) Mask generation losses $\mathcal{L}_{G_m}$, (iv) Bounding box generation losses $\mathcal{L}_{G_b}$, (v) Contrastive loss $\mathcal{L}_{cont}$, (vi) Final cross-entropy classification $\mathcal{L}_{CCE_f}$.}
    \label{fig:scene_designer}
\end{figure*}

\noindent {\bf 2. Sketch-to-Scene Synthesis.}  We hallucinate entire scene layouts from either fully or partially sketched compositions, via a decoder that generates and combines object masks synthesized from learned correlations between object type, appearance and arrangement.  The scene layouts drive texture synthesis to generate photographic content.

\noindent {\bf 3. Fusion of Synthesis and Cross-Domain Search.}  Training a single mixed-task model yields improvements on the state of the art for both scene search and scene synthesis tasks. We also enable a novel creative workflow in which image compositions may be created iteratively; users sketch an initial composition (one or more objects), elements of the resulting images may be incorporated potentially with further sketched objects to drive either further searches or full image synthesis. Fig.~\ref{fig:teaser} demonstrates how our fused model enables iterative design of generative compositions.



\section{Related Work}

{\bf Sketch based image retrieval (SBIR)} has received extensive attention in the past decade. Dictionary learning methods (e.g. bag of visual words) have leveraged wavelet,  edge-let, shape-let and sparse gradient features \cite{Eitz2011,Hu2013} to match sketches to edge structures in images.  With the advent of deep learning, CNNs (convnets) were rapidly adopted for cross-modal representation learning; exploring joint search embeddings for matching structure between sketches and images.  Early approaches learned mappings between edge maps and sketches using contrastive \cite{wang2015sketch3d}, siamese \cite{qi2016} and triplet networks~\cite{Bui2017compact}. Fine-grained SBIR was explored by Yu \etal \cite{yu2016SketchMeShoe} and Sangkloy \etal \cite{sangkloy2016sketchy} who used a three-branch CNN with triplet loss.  Bui \etal learned a cross-domain embedding through triplet loss and partial weight sharing between sketch-image encoders~\cite{Bui2018sketching}.  Later studies followed a natural extension from single object sketches to scenes with multiple objects. Liu \etal~\cite{Liu2020SceneSketcher} recently used a graph encoder with triplet loss to do compositional SBIR, but their results were dependent on known bounding boxes and category labels of photos and sketches. Our method also explores sketched compositions for search, learning a GNN and Transformer hybrid model for both search and synthesis.

{\bf Conditional Image Generation} has been a rapidly developing field since the introduction of Conditional GANs \cite{Mirza2014}. Initial work exploring  class label priors were soon followed by models with image based priors such as Pix2Pix \cite{pix2pix2016} and more recently high definition synthesis models such as Pix2PixHD \cite{wang2018pix2pixHD} and SPADE \cite{park2019SPADE}. Those models learn from pairs of matching samples from each domain, and are capable of mapping semantic layouts to images.

\textbf{Scene Graphs} are a representation for images where individual objects are defined as nodes with the graph edges describe their relationships. They were first used for text-based image retrieval \cite{Johnson2015SceneGraphRetrieval} but were more recently used by Johnson \etal \cite{Johnson2018SceneGraphGeneration} and Ashual \etal \cite{Ashual2019SceneGraphGeneration} as a initial representations to drive semantic layout generation and image synthesis based on those layouts. Our synthesis approach follows a similar principle, where our model learns to generate semantic layouts and a separate GAN (SPADE \cite{park2019SPADE}) synthesizes the final image.

\textbf{Image-synthesis from Sketch} is a challenging problem due to the abstract and  ambiguous nature of sketches. Recent work such as Sketchy-GAN \cite{Chen2018SketchyGAN} and ContextualGAN \cite{Lu2018ContextualGAN} achieved promising results generating images from single-object sketches, but as was shown in \cite{Gao2020SketchyCOCO}, cannot accommodate sketched scenes with multiple objects. Gao \etal \cite{Gao2020SketchyCOCO} were the first to implement image synthesis from sketch scenes and introduced the paired SketchyCOCO dataset. Our approach differs, in that we learn a single mixed-task model for both search and synthesis --- improving quantitatively and in user study synthesis results.

\textbf{Joint image search and synthesis} has been largely unexplored. Early work includes Sketch2Photo~\cite{chen2009sketch2photo} that supports image search using text and sketch queries then compose an output image from objects of interest found in the returned results. This approach uses statistical methods for image composition thus does not yield high quality output. In a more recent work, Sketchformer~\cite{Ribeiro2020Sketchformer} also supports search and synthesis but only for sketch (single-domain). Similarly, Pang \etal \cite{pang2017cross} uses a sketch decoder mainly to assist their cross-domain learning thus have poor synthesized results. Our method learns both a cross-domain embedding for scene sketches and images and also enables image synthesis from composed objects in either domains.


\section{Methodology}

We developed a cross-modal representation learning framework in order to represent images and sketches with multiple objects in a common feature space; this representation is useful to both cross-modal retrieval and generation. Making use of scene graph representations, our framework looks at objects in a scene using a hierarchy based on object correlation.  Fig. \ref{fig:scene_designer} describes the architecture and stages of the representation, which are each summarized below:

\begin{enumerate}
\itemsep-0.2em 
\item{\textbf{Object-level representation (OLR):} Given an input composition, we encode individual objects (which might be sketched or photo-real) to a common representation.  We refer to this embedding, in which each object is independently represented, as the OLR (subsec. \ref{sec:olr}). }
\item{\textbf{Scene Graph (SG)}: is formed using the OLR to encode objects, along with the discrete positional relationships of all object pairs. }
\item{\textbf{Constrained-correlated representation (CCR):}  A graph neural network (GNN) encodes nodes in the Scene Graph (SG) into a continuous representation, encoding object appearances and their pairwise correlations (subsec. \ref{sec:ccr}).} 
\item{\textbf{Freely-correlated representation (FCR):}  The sets of correlated vectors are fed into a Transformer module, where the attention layers allow for free correlation between all objects. The FCR encodes relationships between each object and all other objects in a weighted manner (subsec. \ref{sec:fcr}). This representation is used for synthesis (subsec. \ref{sec:syn}).}

\item{\textbf{Scene representation (SR):} Together with the FCR, the Transformer module computes a separate single-vector latent space to which metric learning is applied to train a search embedding (subsec. \ref{sec:sbir}).} 
\end{enumerate}

We now describe in further detail how each stage of the representation is learned.

\subsection{Object-level Representation (OLR)}
\label{sec:olr}

We begin by independently encoding each object within the input composition to a common feature embedding, regardless of its input modality.  Following contemporary single-object SBIR \cite{Bui2018sketching, sangkloy2016sketchy, yu2016SketchMeShoe} we adopt a deep metric learning approach using a heterogeneous triplet architecture (\ie no shared weights between the anchor (a) and positive/negative (p/n) branches).  Our network comprises a MobileNet \cite{Howard2017Mobile} backbone, terminated with two shared fully connected (fc) layers, the latter yielding the 128-D OLR embedding.  

The MobileNet is pre-trained on ImageNet \cite{ILSVRC15},  and finetuned for cross-domain learning using a combined categorical cross-entropy (CCE) and triplet loss, with the latter defined as:\vspace{-0.4em}
\begin{equation}
\begin{split}
    \mathcal{L}_{tri} \left(a, p, n\right) = \max\{0, m & + |f_{s}(a), f_{i}(p)|_2 \\
    & - |f_{s}(a), f_{i}(n)|_2\} 
\end{split}
\end{equation}\vspace{-0.4em}

\noindent where $f_{s}(.)$ is the MobileNet that encodes sketches, $f_{i}(.)$ the one that encodes photo-real objects cropped from images, each followed by shared-domain MLP (2 fc layers) that encodes the MobileNet`s outputs to the OLR, $m=0.5$ is the margin and $|.|_2$ is the $L_2$ norm.  Hereafter, we use shorthand $f(.)$ to refer to the encoder irrespective of its modality. The OLR is trained using rasterized sketches (a), and objects cropped from the COCO-stuff dataset with  corresponding (p) and differing (n) object class; see subsec.~\ref{sec:datasets} for dataset details.  To aid convergence, the equal-weighted CCE loss $CCE(.)$ is applied to a further fc layer $f_e(.)$ appended to the network:
vspace{-0.4em}
\begin{eqnarray}
    \mathcal{L}_{cce}(a,p,n) &=& \sum_{c \in \{a,p,n\}}CCE(f_e(c),\hat{c}))
\end{eqnarray}
\vspace{-0.4em}

\noindent where $\hat{a}$,$\hat{p}$, $\hat{n}$ are the one-hot vectors encoding the semantic class of each input.
 
\subsection{Constrained-Correlated Representation (CCR)}
\label{sec:ccr}

A scene graph (SG) describes objects and their spatial relationships within the composition. Formally an SG is a graph $\mathcal{G} = (\mathcal{V},\mathcal{E})$ where $\mathcal{V} = \{ o_1, \cdots, o_i, \cdots, o_{\kappa} \}$ is set of nodes representing the $\kappa$ objects, and $(o_i, r_{ij}, o_j) \in \mathcal{E}$ the set of edges encoding  discrete relationships, $r_{ij} \in \mathcal{R}$  (we used ``left of'', ``right of'', ``above'', ``below'', ``contains'', ``inside of'').

We encode the scene graph via a graph neural network (GNN). We represent each node via the OLR as $f(o_i)$, encoding the pose and appearance of the object.  As in \cite{Ashual2019SceneGraphGeneration}, we adopt a learnable embedding $f_r(r_{ij})$ to map relationships $\mathcal{R}$ into dense vectors.  The GNN $g_r(.)$ comprises a sequence of 6 layers $g_r^k(.)$. In each we follow the architecture of \cite{Johnson2018SceneGraphGeneration}, and edges are processed by an fc layer $f_{fc1}^k$:
\vspace{-0.4em}
\begin{equation}
\begin{split}
    <v_i^{0}, r_{ij}^{0}, v_j^{0}> & =  <f(o_i), f_r(r_{ij}), f(o_j)> \\
    <\hat{v}_i^{k}, r_{ij}^{k+1}, \hat{v_j}^{k}> & = f_{fc1}^k(<v_i^{k}, r_{ij}^{k}, v_j^{k}>)
\end{split}
\end{equation}\vspace{-0.4em}

\noindent that updates the vector for each relationship to $r_{ij}^{k+1}$ and builds the set $\hat{v}_i^{k} \in V^k$ of object vectors. A function ($h$) gathers vectors in $\mathcal{V}$ that describe the same object and averages them; finally they are processed by another fc layer to compute the updated $v_i^{k+1} = f_{fc2}^k(h(V^{k}))$ for each object.  These layers are shared between images and sketches and trained end-to-end as part of our model to learn the \textit{constrained-correlated representation} (CCR).


\subsection{Freely-Correlated Representation (FCR) and Scene Representation (SR)}
\label{sec:fcr}

\begin{figure}
    \centering
    \includegraphics[width=0.9\linewidth]{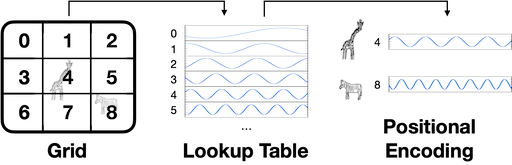}
    \caption{Grid-based Positional Encoding. We adapted the positional encoder from traditional Transformers by attributing one encoding to each block of the gridded scene.}
    \label{fig:pos_encoding}
\end{figure}

We require a final representation that allows for objects to learn from their connection to all others in the scene, without being constrained to the pairwise encoding of the SG.  We model this via attention layers, stacked similarly to a Transformer architecture \cite{Vaswani2017Transformer}. This approach was inspired by the use of LSTMs for SG representation in \cite{Guo2019SceneCaption} in order to accommodate a variable number of objects within a single representation. Transformers offer an attention mechanism to analyze each vector against all others, making them a good candidate for our model; They employ positional encoding so that the sequence order information is kept during processing. Our objects do not have a defined order, but we want their spacial positions to be taken into account. Following \cite{Ashual2019SceneGraphGeneration}, we break the scene into a 5x5 grid and attribute a position $p \in {0, 1, ..., 24}$ to each object based on where its center falls on the grid. This $p$ is used to compute its positional encoding (see Fig. \ref{fig:pos_encoding}). Note that Ashual \etal concatenated this $p$ to the object embeddings processed by the GNN; we took our approach because the vectors processed by our GNN are dynamic and because it aligns better with the Transformer architecture.

We specify our Transformer $t(.) = Z$ with 3 layers, and 16 attention heads on each. The input is the CCR $g_r(f(o_i))$ for each object in the image and one empty vector $\vec{0}$. In each layer $t^i(.)$ the model computes the attention weights that relate each vector to all the vectors in the sequence:\vspace{-0.4em}
\begin{equation}
\begin{split}
    Z^{0} &= s\left(F_q^0(Q+E)F_k^0(Q+E)^T\right)F_v^0(Q+E) \\
    Z^{i+1} &= s\left(F_q^i(Z^i)F_k^i(Z^i)^T\right)F_v^i(Z^i),
\end{split}
\end{equation}\vspace{-0.4em}

\noindent where multiplications are matrix-based, $s$ is the softmax function, $F_q^i$, $F_k^i$, $F_v^i$ are fc layers, $Q$ is a matrix made by stacking $\vec{0}$ and $g_r(f(o_i))$ CCR vectors and $E$ is our positional encodings $f_p(p)$, stacked in the same fashion as $Q$. In the output, the position that contained $\vec{0}$ has the SR while the other ones contain the FCR for each object. 

\subsection{Semantic Layout Generation}
\label{sec:syn}

\begin{figure}
    \centering
    \includegraphics[width=\linewidth,height=2cm]{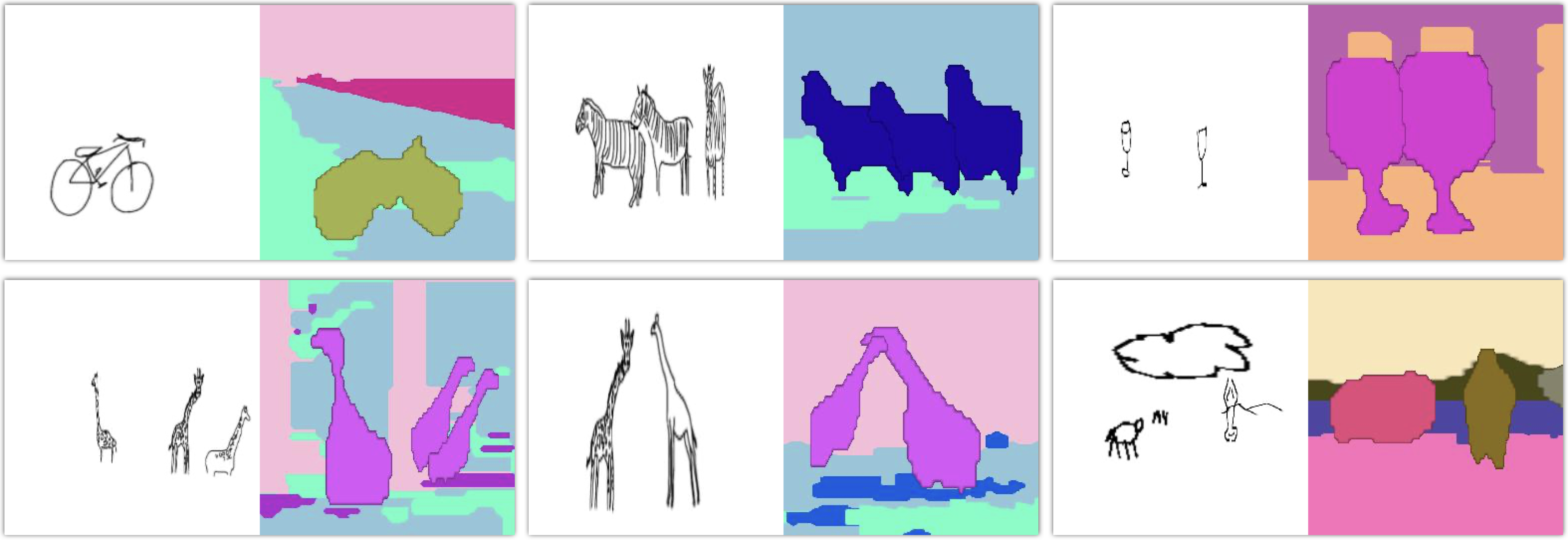}
    \caption{Example layouts synthesized by Scene Designer before they are transformed to images by the SPADE \cite{park2019SPADE} generator. First 2 columns show SketchyCOCO scenes and third column shows QuickDrawCOCO-92c scenes.}
    \label{fig:gen_layouts}
\end{figure}

The learned FCR is used to synthesize a semantic scene layout (c.f. Fig.~\ref{fig:gen_layouts}) that is used by our image generator. The layout is made from bounding boxes and masks for each object. We train two generators, one for masks and one for bounding boxes. The box generator is trained using Generalized Intersection-over-Union (GIOU) \cite{Rezatofighi2019GIOU} and mean-squared error (MSE). The generator itself is a 2-layer MLP. The mask generator is trained within a GAN \cite{Goodfellow14} framework, using a conditional setup \cite{Mirza2014} with LSGAN \cite{Mao2016LSGAN} losses plus MSE and Feature Matching to regularize the adversarial training. Additionally, another CCE classification loss $\mathcal{L}_{CCE_f}$ is applied to the FCR to encourage it to remain semantic.

Taking $b$ as ground truth bounding box, $\hat{x}_i$ as the FCR based on image input and $\hat{x}_s$ as the FCR based on sketch input, our generation losses are, for bounding boxes:\vspace{-0.4em}
\begin{equation}
    \mathcal{L}_{G_{b}}(x) = \lambda_1\mathcal{L}_{GIOU}(b, G_{b}(\hat{x})) 
    +\lambda_2|b, G_{b}(\hat{x})|_2
\end{equation}\vspace{-0.4em}

\noindent for each object on each image, applied to both $\hat{x}_i$ and $\hat{x}_s$, with $\lambda_1=\lambda_2=10$. Using a similar notation, take our ground truth masks as $m_g$, and the object labels as $y \in C$, where $C$ is the set of object classes. The loss for the mask generator is:\vspace{-0.4em}
\begin{equation}
\begin{split}
    \mathcal{L}_{G_{m}}(\hat{x}) &=\lambda_3|m_g, G_{m}(x)|_2 
    +\lambda_4|D(G_{m}(\hat{x}), y), 1|_2 \\
    &+\lambda_5\mathcal{L}_{FM}(\hat{D}(m_g, y), \hat{D}(G_{m}(x), y))
\end{split}
\end{equation}\vspace{-0.4em}

\noindent with $D$ as the discriminator and also applied to both $\hat{x}_i$ and $\hat{x}_s$. The weights are $\lambda_3=10$, $\lambda_4=0.25$, $\lambda_5=10$. The $\mathcal{L}_{FM}(.)$ is the feature matching loss, the L1 difference in the activations of $D$ ($\hat{D}(.)$). $D$ itself is trained with the adversarial loss in classic GAN fashion. Those losses back-propagate through all representation levels.


Finally, we turn the masks and boxes into a semantic layout. In this object-only layout, objects are placed ordered by their size, so that bigger objects are behind smaller ones (after \cite{Ashual2019SceneGraphGeneration}); then it may be combined with a layout for the background (\eg selected from the top search results or coarsely drawn).  Ultimately the final layout is passed through a SotA SPADE model \cite{park2019SPADE} to synthesize an image.

\subsection{Compositional Sketch-based Image Retrieval}
\label{sec:sbir}

For scene retrieval, we learn a single \textit{Scene Representation} (SR) as a search embedding in which the similarity of images and the sketched input may be measured. This is an additional latent vector computed by the Transformer, that has metric properties encouraged through an adaptation of the contrastive loss.

\begin{figure}[b!]
    \centering
    \includegraphics[width=\linewidth]{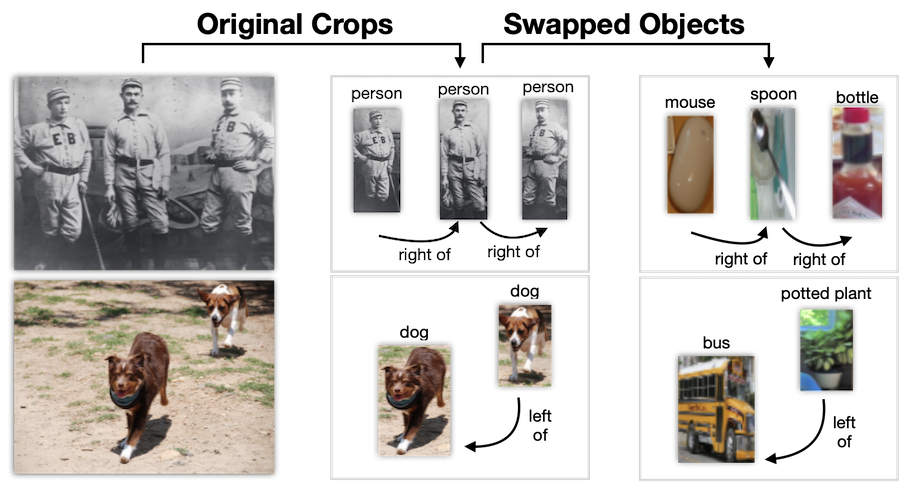}
    \caption{During contrastive training of the SR, negatives are synthesized by swapping objects in the scene for others of differing class. This helps the model prioritize class over structure. First column shows the original image, the second the individual cropped objects, and the third the swapped objects.}
    \label{fig:shuffled_objs}
\end{figure}

 During contrastive training, we sample half of our negatives randomly from other images of scenes in the training set.  The other half of the negatives are synthesized by swapping objects of different class into the positions of the objects in the positive image (Fig.~\ref{fig:shuffled_objs}).  By including such a class-swapped version of the image as a negative, the model is encouraged to discriminate change in object class to a greater degree than changes in object positions. Our  contrastive loss, adapted to incorporate these synthetic negatives, is:\vspace{-0.4em}
\begin{equation}
\begin{split}
    \mathcal{L}_{cont}(x_{s}, x_{i}, x_{sn}) &= \dfrac{1}{2}(Y)(\mathcal{D}(x_{s}, x_{i})) \\
    &+ \dfrac{1}{4}(1 - Y)(1 - \mathcal{D}(x_{s}, x_{i})) \\
    &- \dfrac{1}{4}\mathcal{D}(x_{s}, x_{sn}),
\end{split}
\end{equation} \vspace{-0.4em}

\noindent where $\mathcal{D}$ computes $|.|_2$  between all vectors on one set and all vectors on another set, $x_{s}$ and $x_{i}$ are respectively the sketch-based and image-based SR, $Y$ is a label indicating if a pair of vectors are from the same scene. Finally, $x_{sn}$ is the SR of the synthesized scenes with swapped objects. 

\subsection{Training Stages}

Training is performed in three stages. First the OLR is trained independent of the rest of the model, using the dual triplet and categorical cross-entropy loss, for 100K iterations, on single-object sketches and object crops.  In the second stage we use our novel soft-paired sketch and image scenes dataset (see Sec. \ref{sec:datasets}) to train entire model end-to-end for a further 120K iterations. Finally, we finetune the model with a hard-paired dataset, training for a further 5K iterations. Training is via the ADAM \cite{Kingma2015ADAM} optimizer, with $\beta_1=0.5$ and $\epsilon=1e-9$; learning rate (lr) is $lr=10^{-4}$ for the first two stages and $lr=10^{-5}$ for the finetuning stage. The mask discriminator follows the \textit{Two Time-Scale Update Rule} (TTUR) \cite{Heusel2017TTUR} with $lr_D=4lr$.

\section{Experiments and Discussion}
\label{sec:exp}

We evaluate the performance of Scene Designer for both compositional sketch search (SBIR) and synthesis, contrasting performance against contemporary baselines for both tasks.  We explore the efficacy of our model for both search and synthesis from sketch, image and mixed domain compositions.  For SBIR we compare against four baselines: a scene-level technique (SceneSketcher \cite{Liu2020SceneSketcher}) and three single-object techniques. Of these, two are fine-grained SBIR models (Sketchy \cite{sangkloy2016sketchy}, and Sketch-me-that-shoe \cite{yu2016SketchMeShoe}),  and one is coarse-grained (Multi-stage Learning or MSL \cite{Bui2018sketching}).  For scene synthesis we compare against the sketch driven method of Gao \etal \cite{Gao2020SketchyCOCO} (proposed alongside their SketchyCOCO dataset) and the method of Ashual \etal \cite{Ashual2019SceneGraphGeneration} that accepts semantic scene graphs (spatial arrangements of keywords) as input.

\subsection{Datasets}
\label{sec:datasets}

We require paired data; sketched compositions and corresponding images of the scenes depicted by those sketches.  

\noindent{\bf SketchyCOCO} is a recent dataset of 14K such pairs, proposed for sketch based scene synthesis \cite{Gao2020SketchyCOCO}.  SketchyCOCO samples scenes and associated annotations (bounding boxes, masks) from the COCO-stuff \cite{caesar2018COCOstuff} images dataset, augmenting these with sketched depictions of those scenes.  Approximately 4K of SketchyCOCO images contain object instances (the remainder are solely background material; `stuff').  Only 14 classes (14c) of objects are annotated.   To mitigate overfitting, we use the SketchyCOCO training partition (4,060 samples) only for fine-tuning our model.  We include the test partition in our evaluation and split it per experiment requirements: retrieval eval. uses 233 samples as \cite{Liu2020SceneSketcher} needs scenes with 2 or more objects; generation tests needs to use the overlap between test sets used in each of the compared models: 137 samples. We also emulate  \cite{Liu2020SceneSketcher} in constructing an `Extended SketchyCOCO' search corpus by holding out 5K random images from the COCO-stuff training set and adding these to the SketchyCOCO test set as distractors.

\begin{figure}[t!]
    \centering
    \includegraphics[width=\linewidth, height=2.8cm]{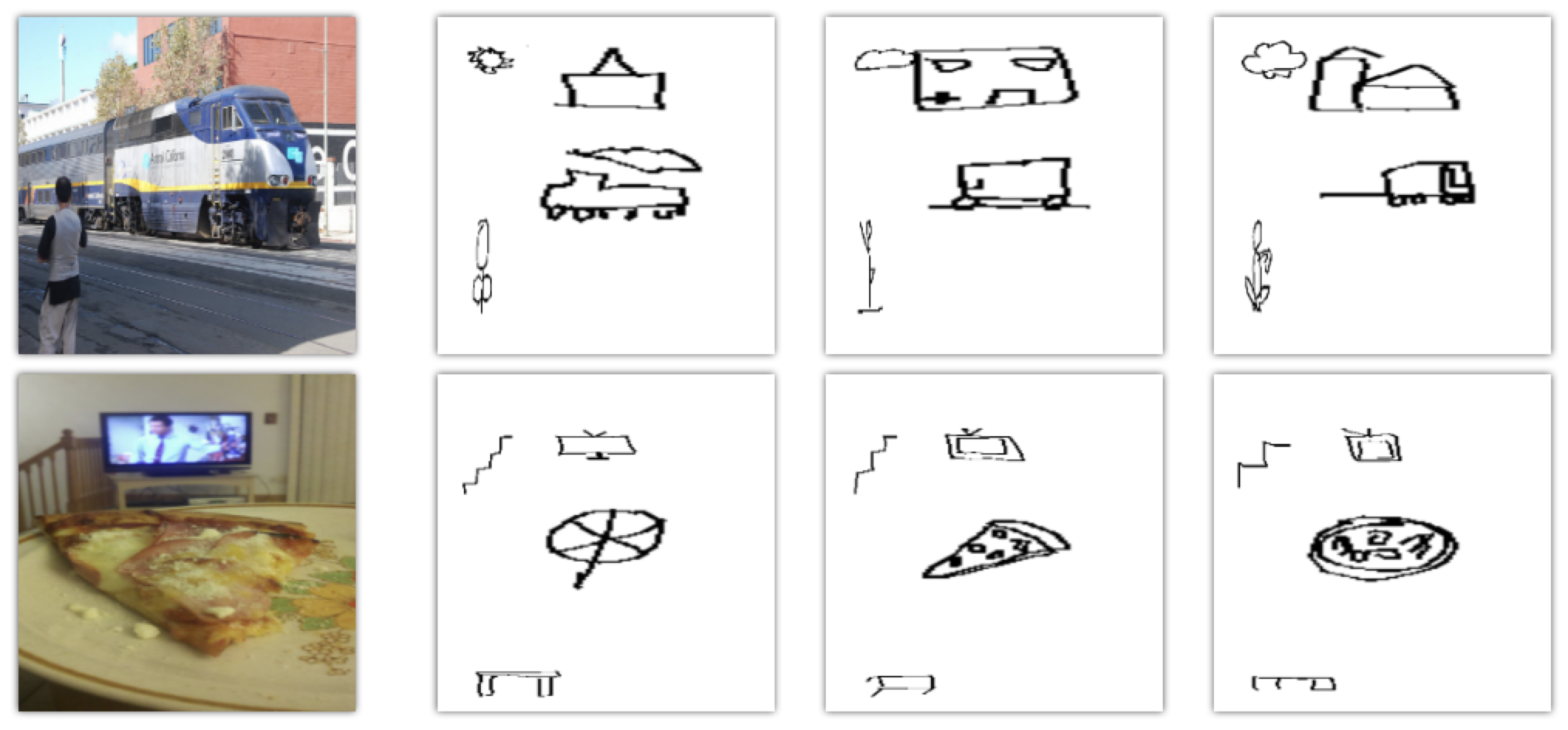}
    \caption{Sample paired data from the proposed QuickDrawCOCO-92c dataset. Each row shows one COCO-stuff \cite{caesar2018COCOstuff} image and three examples of synthetic sketched compositions matching that image derived from QuickDraw \cite{QD50}.}
    \label{fig:qdcoco92}
\end{figure}

\noindent{\bf QuickDrawCOCO-92c} is a novel dataset we propose, composing scenes with sketches from QuickDraw, the largest public sketch dataset (50M) \cite{QD50}.  Inspired by SketchyCOCO, we similarly leverage scenes from COCO-stuff, taking advantage of the class overlap (92c) with QuickDraw.  We synthesize sketch scenes by compositing sketches from QuickDraw onto a canvas, selecting a random sketch within the class corresponding to each object in a COCO-stuff image (Fig. \ref{fig:qdcoco92}). Our dataset is soft-paired, in that the sketches for each object do not necessarily match pose and appearance with the image, only the class label. We use QuickDrawCOCO-92c for all training stages except for the final finetuning stage where we combine samples from both SketchyCOCO and QuickDrawCOCO-92c.  We separately fuse the  training and test partitions of the QuickDraw and COCO-stuff datasets to partition QuickDrawCOCO-92c into 111,112 training, 2,788 test and 1,907 validation scenes. This dataset is 6x more category-diverse and has 8x more samples than SketchyCOCO.

\subsection{Evaluating Retrieval}

\begin{figure}[t]
    \centering
    \includegraphics[width=\linewidth, height=6.5cm]{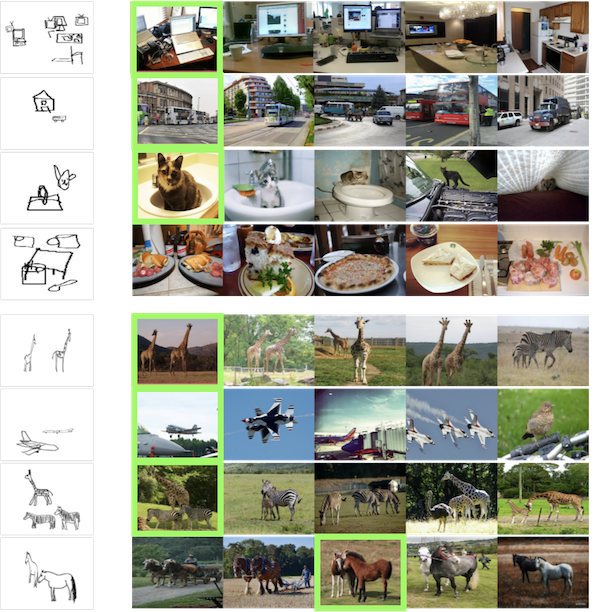}
    \caption{Retrieval results using the QuickDrawCOCO (first block) and SketchyCOCO sets (second). The results often contain the exact match and are contextually consistent, such as in the last row of QuickDrawCOCO, where the context of ``table-spoon-mug-etc'' retrieves sets of ``meals''.  Relevant results in green; final row of each block represents a failure case.}
    \label{fig:qdcoco-scoco-sbir}
\end{figure}

We adopt the evaluation proposed in SceneSketcher \cite{Liu2020SceneSketcher}, the only prior technique exploring SBIR for scenes; retrieved images are relevant only if they match the sketched scene exactly. Models are evaluated on SketchyCOCO, QuickDrawCOCO-92c and `Extended SketchyCOCO'.

\textbf{Compositional search.} Tab.~\ref{tab:scoco-qdcoco-SSBIR} reports recall at rank $k=\{1,5,10\}$ (R@k) for our proposed search embedding for sketched scenes, that of SceneSketcher \cite{Liu2020SceneSketcher}, and three single-object SBIR baselines: Sketch-me-that-shoe \cite{yu2016SketchMeShoe}; Sketchy \cite{sangkloy2016sketchy}; and MSL \cite{Bui2018sketching}.  We report R@k over SketchyCOCO and Extended SketchyCOCO for all methods; figures for SceneSketcher and Sketch-me-that-shoe are taken from Liu \etal \cite{Liu2020SceneSketcher}.  We are unable to report values for these two models over our QuickDrawCOCO-92c test data, due to the lack of public models/code for these methods.  QuickDrawCOCO-92c is a more challenging query set due to its increased size and the more messy / abstract sketches within QuickDraw. We present our results alongside public Sketchy \cite{sangkloy2016sketchy} and MSL \cite{Bui2018sketching} models  (final column, Tab.~\ref{tab:scoco-qdcoco-SSBIR}).

Our model more than doubles previous SotA recall results and greatly improves other metrics. The penultimate row of Tab.~\ref{tab:scoco-qdcoco-SSBIR} reports performance our method using mixed-domain queries, where we substitute half of the sketched objects for their image cropped counterparts, which yields an even higher result likely due to half of the objects sharing the same domain as the test corpus.  The final row reports on our model trained on QuickDrawCOCO-92c only; prior to finetuning on SketchyCOCO training set.  This demonstrates that our model generalizes sufficiently to beat the SotA on SketchyCOCO without explicit training on it. Visual results for both datasets are in Fig. \ref{fig:qdcoco-scoco-sbir}.

\textbf{Single object search.} We evaluate the performance of our model at single-object retrieval, baselining against both the Sketchy and MSL models on the SketchyCOCO dataset.  We sample 120 queries randomly from SketchyCOCO containing only one object.  In contrast to composition search, many relevant results exist in the SketchyCOCO dataset for a given sketched object.  We consider a result relevant if both its category and pose match the sketched query. Since no ground-truth annotation with these criteria is available for SketchyCOCO, we crowd-source per-query annotation via Mechanical Turk (MTurk) for the top-$k$ ($k$=15) results and compute both recall@$k$ and precision@$k$  in Tab.~\ref{tab:mturk-sbir}. For each ranked result we ask 5 MTurk participants to indicate relevance (or not), and filter on consensus level.  Our approach significantly outperforms both baselines at all consensus levels.  The sensitivity to object orientation and pose underlines the utility of sketch over labeled boxes (per \cite{Ashual2019SceneGraphGeneration}) to specify the appearance of the desired object.

\begin{table*}
\small
\centering
\begin{tabular}{lccc|ccc|ccc}
\multirow{2}{*}{\ttfamily Method} & \multicolumn{3}{c|}{\ttfamily SketchyCOCO} & \multicolumn{3}{c|}{\ttfamily Ext. SCOCO} & \multicolumn{3}{c}{\ttfamily QuickDrawCOCO} \\
& \ttfamily r@1 & \ttfamily r@5 & \ttfamily r@10 & \ttfamily r@1 & \ttfamily r@5 & \ttfamily r@10 & \ttfamily r@1 & \ttfamily r@5 & \ttfamily r@10 \\ 
\hline \hline
Ours & \textbf{75.53} & \textbf{96.56} & \textbf{99.14} & \textbf{66.09} & \textbf{90.55} & \textbf{96.13} & \textbf{62.15} & \textbf{78.90} & \textbf{85.07} \\
Sketch-me-that-shoe \cite{yu2016SketchMeShoe} & 06.19 & 17.15 & 32.86 & \textless0.01 & \textless0.01 & 01.90 & - & - & - \\
Sketchy \cite{sangkloy2016sketchy} & 03.43 & 09.87 & 15.87 & \textless0.01 & 00.85 & 00.85 & 00.07 & 00.35 & 00.60 \\ 
MSL \cite{Bui2018sketching} & 09.44 & 28.75 & 43.34 & 05.15 & 15.87 & 17.59 & 00.10 & 00.53 & 01.07 \\
SceneSketcher \cite{Liu2020SceneSketcher} & 31.91 & 66.67 & 86.19 & 12.38 & 26.67 & 38.10 & - & - & - \\ \hline
Ours (Mixed-domain) & 89.69 & 99.57 & 100.0 & 87.12 & 96.56 & 98.71 & 93.22 & 97.37 & 98.23 \\
Ours (Before finetuning)  & 45.49 & 80.25 & 93.56 & 15.45 & 41.20 & 50.21 & 64.27 & 81.95 & 87.23 \\ \hline
\end{tabular}
\caption{Recall@K metrics on the SketchyCOCO dataset, its extended version and the QuickDrawCOCO-92c set. Our model more than doubles the previous state-of-the-art recall@1 metric. Additionally, we show our performance when using mixed-domain compositions and also with the model trained only with QuickDrawCOCO-92c, before finetuning on SketchyCOCO.}
\label{tab:scoco-qdcoco-SSBIR}
\end{table*}

\begin{table}[ht]
\small
\centering
\begin{adjustbox}{width=1\linewidth}
\begin{tabular}{lccc|ccc}
\multirow{2}{*}{\ttfamily Method} & \multicolumn{3}{c|}{\ttfamily Precision@1} & \multicolumn{3}{c}{\ttfamily Precision@15} \\ 
& \ttfamily C2 & \ttfamily  C3 & \ttfamily  C4 & \ttfamily  C2 & \ttfamily  C3 & \ttfamily  C4 \\ \hline \hline

Ours & \textbf{66.12} & \textbf{52.07} & \textbf{37.19} & \textbf{22.98} & \textbf{16.14} & \textbf{10.30} \\
MSL \cite{Bui2018sketching} & 16.53 & 14.05 & 09.09 & 10.41 & 07.00 & 04.41 \\
Sketchy \cite{sangkloy2016sketchy} & 35.54 & 23.97 & 18.18 & 17.36 & 11.79 & 07.77 \\

\hline
& \multicolumn{3}{c|}{\ttfamily Recall@1} & \multicolumn{3}{c}{\ttfamily Recall@15} \\
& \ttfamily C2 & \ttfamily  C3 & \ttfamily  C4 & \ttfamily  C2 & \ttfamily  C3 & \ttfamily  C4 \\
\hline \hline

Ours & \textbf{66.12} & \textbf{52.07} & \textbf{37.19} & \textbf{96.69} & \textbf{90.91} & \textbf{71.90} \\
MSL \cite{Bui2018sketching} & 16.53 & 14.05 & 09.09 & 84.30 & 72.73 & 56.20 \\
Sketchy \cite{sangkloy2016sketchy} & 35.54 & 23.97 & 18.18 & 85.95 & 80.17 & 65.29\\

\hline
\end{tabular}
\end{adjustbox}
\caption{MTurk sketch search results for the proposed and baseline methods on the SketchyCOCO dataset. Precision@k and Recall@k are presented at different (C)oncensus thresholds.}
\label{tab:mturk-sbir}
\end{table}

\subsection{Evaluating Scene Synthesis}

\begin{figure*}[t!]
    \centering
    \includegraphics[width=\linewidth,height=4.2cm]{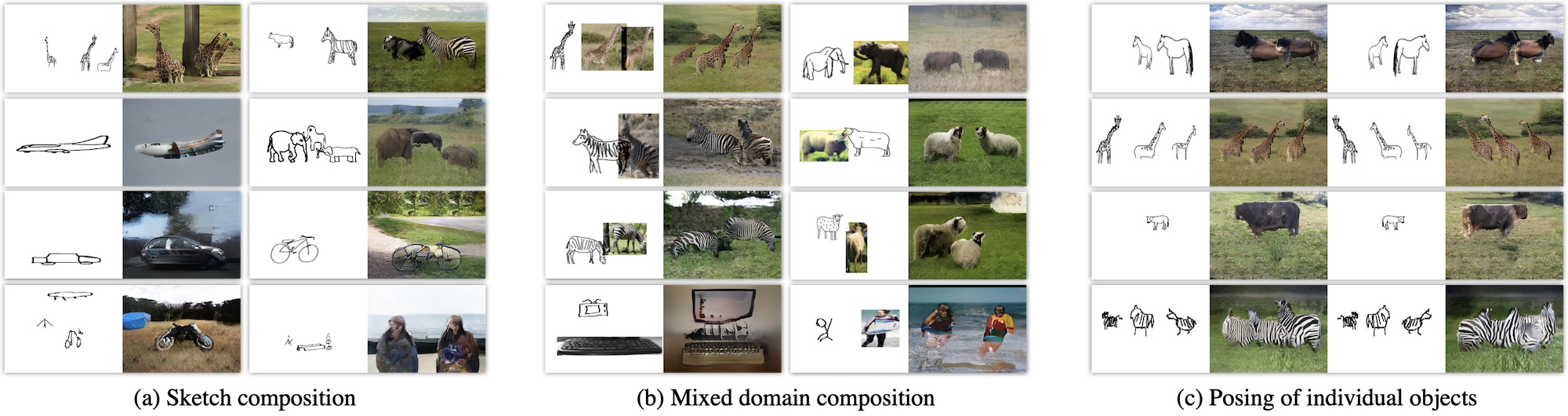}
    \caption{Generating images with Scene Designer. The first 3 rows show compositions and sketches from SketchyCOCO whilst the last one shows those from QuickDrawCOCO-92c. There are three features shown: (a) sketch composition. (b) compositions that mix image and sketched objects. (c) individual object posing, where we flipped each object individually and compare the generations.}
    \label{fig:scoco_generation}
\end{figure*}

We evaluate the performance of our proposed model at image synthesis (see Fig.\ref{fig:scoco_generation}) using SketchyCOCO, comparing  against that dataset's accompanying public model \cite{Gao2020SketchyCOCO}, and against the public model of Ashual \etal \cite{Ashual2019SceneGraphGeneration} which synthesizes images from spatial word maps \ie implicit scene graphs. We synthesize from the sketch, and use the paired image as ground truth for evaluating the fidelity of the output.  For Ashual \etal, we create an input scene graph based on object classes and bounding boxes.

\textbf{Objective Metrics.} We compute Fr\'{e}chet Inception Distance (FID) \cite{Heusel2017TTUR} and Object Classification Accuracy \cite{Ashual2019SceneGraphGeneration} to compare the synthesized and ground truth images. A lower FID value indicates that the tested image distribution is closer the original and is more realistic. The Accuracy values come from finetuning a ResNet-101 classifier on crops from COCO and applying it to foreground objects cropped from the generated images. 

Tab.~\ref{tab:scoco-gen} shows our model to score higher than both baselines on both FID and Accuracy. We have included results applying SPADE to the ground-truth layout, which serves as an upper bound generation result for our method. Results from mixed-domain and images-only compositions, which are a unique capability of our model, are also included.  These show the value in applying Scene Designer to incrementally build up a composition as part of an interactive creative process.

\textbf{User Perceptual Study.} We assess the quality of our synthesized images following the subjective evaluation methodology proposed by Gao \etal \cite{Gao2020SketchyCOCO}. Specifically we score how `realistic' each image is, and how `faithful' each synthesized image is in representing the spatial object arrangement of the input  scene (in the case of \cite{Ashual2019SceneGraphGeneration}, spatial arrangement of labeled boxes).
User opinion on `faithfulness' is scored on a scale 1 (`very dissatisfied') to 4 (`very satisfied') using  crowd-sourced annotations collected via MTurk.    For realism, images synthesized by each method are presented to participants, inviting selection of the most realistic.  In both cases each task is annotated by 4 unique participants, and results tabulated for differing levels of participant consensus.  We consider only results where 2 or more MTurkers agree \ie there was consensus. Tab.~\ref{tab:human-eval} indicates our method is preferred for content faithfulness and realism, although faithfulness scores are low in general (circa 2.5 on the scale) this is in line with reported figures for other methods e.g. 1.57 for \cite{Ashual2019SceneGraphGeneration} as shown in \cite{Gao2020SketchyCOCO}.

\begin{table*}
\centering
\small
\begin{tabular}{lccc|ccc|cccccc}
\multirow{3}{*}{\ttfamily Method}           & \multicolumn{6}{c|}{\ttfamily SketchyCOCO}                     & \multicolumn{6}{c}{\ttfamily QuickDrawCOCO}                                \\
                            & \multicolumn{3}{c}{\ttfamily Realism} & \multicolumn{3}{c|}{\ttfamily Faithfulness} & \multicolumn{3}{c}{\ttfamily Realism}       & \multicolumn{3}{c}{\ttfamily Faithfulness} \\ 
& \ttfamily C2 & \ttfamily  C3 & \ttfamily  C4 & \ttfamily  C2 & \ttfamily  C3 & \ttfamily  C4 & \ttfamily  C2 & \ttfamily  C3 & \multicolumn{1}{c|}{\ttfamily C4} & \ttfamily C2 & \ttfamily C3 & \ttfamily C4 \\ \hline \hline
Ours & \textbf{64.17} & \textbf{72.73} & \textbf{82.61} & 2.67 & 2.50 & \textbf{2.16} & \textbf{62.77} & \textbf{63.64} & \multicolumn{1}{l|}{\textbf{75.68}} & \textbf{2.66} & 2.56 & 2.58 \\ 

Ashual \etal \cite{Ashual2019SceneGraphGeneration} & 13.33 & 13.64 & 04.35 & 2.56 & 2.34 & 1.90 & 37.23 & 36.36 & \multicolumn{1}{l|}{24.32} & 2.61 & \textbf{2.58} & \textbf{2.88} \\

SketchyCOCO \cite{Gao2020SketchyCOCO} & 22.50 & 13.64 & 13.04 & \textbf{2.79} & \textbf{2.65} & 2.04 & - & - & \multicolumn{1}{l|}{-} & - & - & - \\

\hline
\end{tabular}
\caption{User perceptual study (via MTurk) evaluating the generated images. Realness is a comparative score between the models while faithfulness is individually scored per method on continuous scale [1,4].  Results are thresholded for different levels of participant (C)onsensus, from 2 to 4 (out of 4) participant agreements.}
\label{tab:human-eval}
\end{table*}

\begin{table}
\centering
\small
\begin{adjustbox}{width=\linewidth,center}
\begin{tabular}{lll|ll}
\multirow{2}{*}{\ttfamily Method} & \multicolumn{2}{l|}{\ttfamily SketchyCOCO} & \multicolumn{2}{l}{\ttfamily QDCOCO-92c} \\
& \ttfamily FID$\downarrow$ & \ttfamily Acc.$\uparrow$ & \ttfamily FID$\downarrow$ & \ttfamily Acc.$\uparrow$ \\ 
\hline \hline
Ours (sketch-based)  & \textbf{130.87} & \textbf{63.46} & \textbf{76.64} & \textbf{65.47} \\
Ashual \etal \cite{Ashual2019SceneGraphGeneration} & 170.40 & 56.20 & 103.14 & 50.69 \\
SketchyCOCO \cite{Gao2020SketchyCOCO} & 198.17 & 29.14 & - & - \\ 
\hline
Ours (image-based) & 138.82 & 57.36 & 69.08 & 59.52 \\
Ours (mixed-domain)  & 143.24 & 65.74 & 83.13 & 58.53 \\ 
SPADE \cite{park2019SPADE} & 111.25 & 81.21 & 58.39 & 80.95 \\ \hline
\end{tabular}
\end{adjustbox}
\caption{Generation metrics when using samples from SketchyCOCO and QuickDrawCOCO-92c. Both quantitative metrics were computed on the overlapping test set across the three/two models for each dataset (137 samples on the former and 1300 on the later). A lower FID and higher Object Classification Accuracy represent a better result. We've also included the metrics for the SPADE generator, representing an upper bound for our model.}
\label{tab:scoco-gen}
\end{table}

\subsection{Ablation studies}
\label{sec:abbl}

\begin{table}
\small
    \centering
        \begin{tabular}{llll}
            \ttfamily Model Settings & \ttfamily r@1 & \ttfamily r@10 & \ttfamily FID$\downarrow$ \\                \hline
            \hline
            \multicolumn{1}{l}{(A1) W/o Transformer} & \multicolumn{1}{l}{07.20} & \multicolumn{1}{l}{20.62} & 102.79 \\
            
            \multicolumn{1}{l}{(A2) W/o GNN} & \multicolumn{1}{l}{02.69} & \multicolumn{1}{l}{17.61} & 118.04 \\
            
            \multicolumn{1}{l}{(A3) W/o positional encoding} & \multicolumn{1}{l}{14.45} & \multicolumn{1}{l}{45.87} & 122.01 \\
            
            \hline
            \multicolumn{1}{l}{(B1) W/o obj.-level pretraining} & \multicolumn{1}{l}{55.84} & \multicolumn{1}{l}{85.43} & 102.14 \\
            \multicolumn{1}{l}{(B2) W/o synthesized negatives} & \multicolumn{1}{l}{54.12} & \multicolumn{1}{l}{81.16} & 128.58 \\
            
            \hline
            \multicolumn{1}{l}{(C1) No contrastive loss} & \multicolumn{1}{l}{45.48} & \multicolumn{1}{l}{74.67} & 129.88 \\
            
            \multicolumn{1}{l}{(C2) No generation losses} & \multicolumn{1}{l}{55.84}  & \multicolumn{1}{l}{81.34} & 129.98 \\
            \hline
            \multicolumn{1}{l}{Final model} & \multicolumn{1}{l}{\textbf{62.15}} & \multicolumn{1}{l}{\textbf{85.07}} & \textbf{76.64}
            
             \\\hline
        \end{tabular}

    \caption{Ablation Studies, showing Recall@k and FID on QuickDrawCOCO. With each set of models, we want to show that (A) each component is necessary, (B) the training procedure aids performance and (C) the single model can multi-task well.}
    \label{tab:abbl}
\end{table}

We explore the significance of each stage of our proposed architecture and training methodology, comparing SBIR and generation results on QuickDrawCOCO-92c using several ablated variants as shown in Tab.~\ref{tab:abbl}, alongside the QuickDrawCOCO-92c result for the full method. We explore three categories of ablation (A-C).  Category A ablates key features of the proposed architecture.  Category B ablates key training steps. Category C investigates if training a single model for both search and synthesis does not degrade the model's performance at either.

\textbf{(A1) Without Transformer. } Removes the attention modules from the model, aggregating the representations from the GNN using simple addition.

\textbf{(A2) Without GNN. } Substitutes the GNN for an fc layer that bridges the OLR into the Transformer module.

\textbf{(A3) Without Positional Encoding. } Retains the proposed architecture but removes our grid-based positional encoding from the Transformer module.

\textbf{(B1) Without object-level pretraining. } Skips the first stage  of training the OLR.

\textbf{(B2) Without shuffled-objects negatives. } Does not use our synthesized negative samples in the contrastive loss.

\textbf{(C1) No Contrastive Loss. } Does not use contrastive loss. The results on this model are driven by the masks and boxes generation losses and object-level triplet loss only.

\textbf{(C2) No Generation Losses. } By removing the masks and boxes generation losses, this ablated model is only trained with search associated losses.


Removing either the GNN (A2) or the grid-based positional encoding (A3) degrades both tasks' performance significantly while the Transformer (A1) is shown to be helpful for generation (improves FID by 25\%), and crucial for search. 

Our synthesized negative examples (B2) improve search by 8\% and are essential for generation to work. This follows the result of (C1) where removing the Contrastive Loss also diminishes generation performance; given that both of those are associated with search but also impact generation show that the strength of our multi-task model lies on the synergy between those tasks. 

When we then look at (C2) that hypothesis is further proven, as removing the synthesis-related losses impact the search results as well with a drop of 6\% in recall@1. Training for both tasks does not decrease performance, but boosts it meaningfully. We conclude that learning general representations requires the model to retain both visual and semantic information, which we achieve by mixing generation and feature learning losses.


\section{Conclusion}
\vspace{-0.15cm}

We introduced Scene  Designer;  a  single  unified model for searching and generating images using free-hand sketches of scenes.  We developed a mixed-task learning framework with three levels of representation (OLR, CCR, FCR) using a hybrid GNN-Transformer architecture. Scene Designer learns to embed sketched and photographic scenes into a common space, producing latent representations for both synthesizing layouts from sketched scenes and for measuring compositional similarity between sketch and image scenes.  The model is trained via an expanded dataset of sketch compositions and corresponding images (QuickDrawCOCO-92c); a secondary contribution of our work. 
We show that the combination of feature learning and generation losses aided by our novel take on contrastive learning for scenes are responsible for obtaining SotA performance at scene search and synthesis tasks. The ability to sketch an initial composition, and incorporate components of results into hybrid queries for synthesis (or further searches) creates a novel mechanism for interactively constructing scenes from digital image collections.  Further work could explore this interactive model in creative practice, and perhaps explore how other facets of generative artwork (\eg neural style transfer) might be incorporated into the framework for example to enable fine-grained control over the appearance of objects within the composition.  

\vspace{-0.25cm}\section*{Acknowledgments}\vspace{-0.1cm}
This work was supported by FAPESP (grants 2017/22366-8, 2019/02808-1, 2019/07316-0), CNPq Fellowship (304266/2020-5), and a charitable donation from Adobe Inc.

{\small
\bibliographystyle{unsrt}
\bibliography{egpaper_for_review}
}

\end{document}